\documentclass[]{bytedance_seed}
\usepackage{helvet}

\usepackage{amsmath} 
\usepackage{natbib}
\usepackage{graphicx}
\usepackage{subcaption}

\usepackage[toc,page,header]{appendix}
\usepackage[utf8]{inputenc} 
\usepackage[T1]{fontenc}    
\usepackage{hyperref}       
\usepackage{url}            
\usepackage{booktabs}       
\usepackage{lmodern}        
\usepackage{amsfonts}       
\usepackage{nicefrac}       
\usepackage{microtype}      
\usepackage{wrapfig}

\usepackage{amssymb}  
\usepackage{fontawesome}  
\usepackage{url}  

\usepackage{minitoc}

\usepackage{enumitem}
\usepackage{tablefootnote}
\usepackage{dsfont}
\usepackage[flushleft]{threeparttable}
\usepackage{multirow}
\usepackage[svgnames]{xcolor}
\usepackage[most]{tcolorbox}
\usepackage{caption}
\usepackage{subcaption}

\usepackage{amssymb}
\usepackage{bbding}
\usepackage{float}

\usepackage{pgfplots}
\pgfplotsset{compat=1.18}

\usepackage[table]{xcolor} 
\usepackage{colortbl}      

\usepackage{booktabs}
\usepackage{array}
\usepackage{etoolbox}

\definecolor{lightblue}{RGB}{200, 230, 255}  
\definecolor{headerblue}{RGB}{150, 200, 255} 

\definecolor{Gray}{gray}{0.8}

\newcommand*{\method}{Adv-GRPO}
\usepackage{multirow}

\title{The Image as Its Own Reward: 
Reinforcement Learning with Adversarial Reward for Image Generation}
\author[1]{Weijia Mao}
\author[2\dagger]{Hao Chen}
\author[2]{Zhenheng Yang}
\author[1\dagger]{Mike Zheng Shou}

\affiliation[1]{Show Lab, National University of Singapore}
\affiliation[2]{ByteDance}

\contribution[\dagger]{Corresponding authors}

\abstract{
\begin{abstract}

A reliable reward function is essential for reinforcement learning (RL) in image generation. Most current RL approaches depend on pre-trained preference models that output scalar rewards to approximate human preferences. However, these rewards often fail to capture human perception and are vulnerable to reward hacking, where higher scores do not correspond to better images. To address this, we introduce \textbf{Adv-GRPO}, an RL framework with an adversarial reward that iteratively updates both the reward model and the generator. The reward model is supervised using reference images as positive samples and can largely avoid being hacked. Unlike KL regularization that constrains parameter updates, our learned reward directly guides the generator through its visual outputs, leading to higher-quality images. Moreover, while optimizing existing reward functions can alleviate reward hacking, their inherent biases remain. For instance, PickScore may degrade image quality, whereas OCR-based rewards often reduce aesthetic fidelity.
To address this, we take \textbf{the image itself as a reward}, using reference images and vision foundation models (e.g., DINO) to provide rich visual rewards. These dense visual signals, instead of a single scalar, lead to consistent gains across image quality, aesthetics, and task-specific metrics. Finally, we show that combining reference samples with foundation-model rewards enables distribution transfer and flexible style customization. In human evaluation, our method outperforms Flow-GRPO and SD3, achieving 70.0\% and 72.4\% win rates in image quality and aesthetics, respectively. Code and models have been released.

\end{abstract}
}

\date{\today}
\correspondence{
Hao Chen at \email{haochen.umd@gmail.com},
Mike Zheng Shou at \email{mike.zheng.shou@gmail.com}, 
}

\checkdata[Project Page]{\url{https://showlab.github.io/Adv-GRPO/}}

\begin{document}
\maketitle



\begin{figure*}[t]
    \centering
    \includegraphics[width=\textwidth]{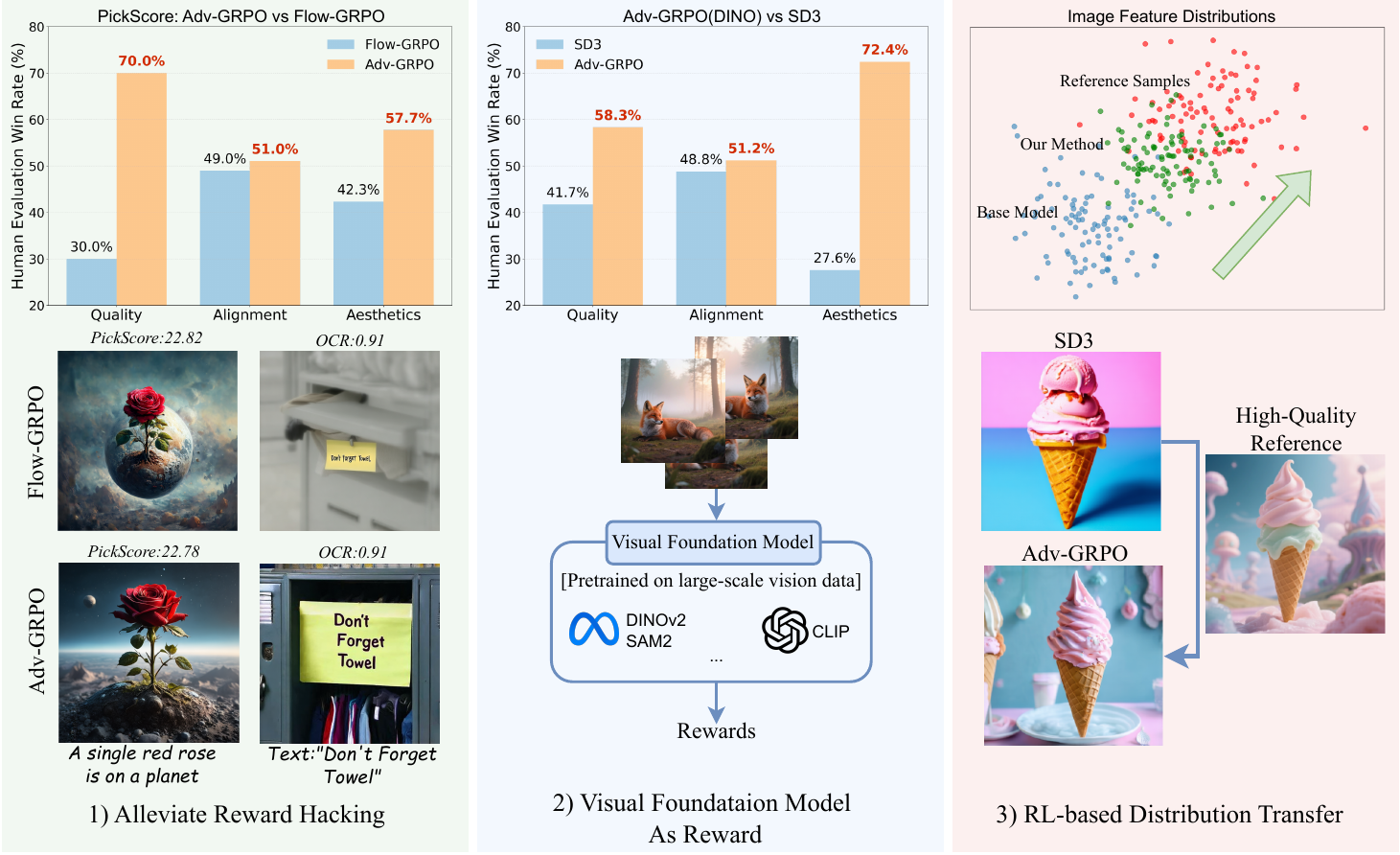}
     \caption{
        \textbf{Overview of our approach.}
        Our method \method{} improves text-to-image (T2I) generation in three ways: 
        \textbf{1) Alleviate Reward Hacking}, achieving higher perceptual quality while maintaining comparable benchmark performance (e.g., PickScore, OCR), as shown in the top-left human evaluation panel; 
        \textbf{2) Visual Foundation Model as Reward}, leveraging visual foundation models (e.g., DINO) for rich visual priors, leading to overall improvements as shown in middle-top human evaluation results; 
        \textbf{3) RL-based Distribution Transfer}, enabling style customization by aligning generations with reference domains.
    }
    \vspace{-15pt}
    \label{fig:teaser}
\end{figure*}

\section{Introduction}
Recently, online reinforcement learning (RL) has attracted increasing attention in large language models (LLMs)~\cite{deepseek_r1,grpo,gspo} and multimodal large language models (MLLMs)~\cite{grpo_vl,unirl,minio3,deepeyes,dr_grpo,pixel_reasoner}. In particular, the Group Relative Policy Optimization (GRPO)~\cite{grpo} algorithm, introduced by DeepSeek-R1~\cite{deepseek_r1}, has proven effective for aligning model behavior with reward signals in these domains. Motivated by these advances, several studies~\cite{flowcps,flowgrpo,dancegrpo,DRaFT,imagereward} have applied online RL to text-to-image (T2I) generation with diffusion models. For example, DanceGRPO~\cite{dancegrpo} and Flow-GRPO~\cite{flowgrpo} demonstrate that reward-driven optimization can improve performance when suitable reward models are provided.

\begin{wrapfigure}{r}{0.55\linewidth}  
    \vspace{-10pt}    
    \centering
    \includegraphics[width=\linewidth]{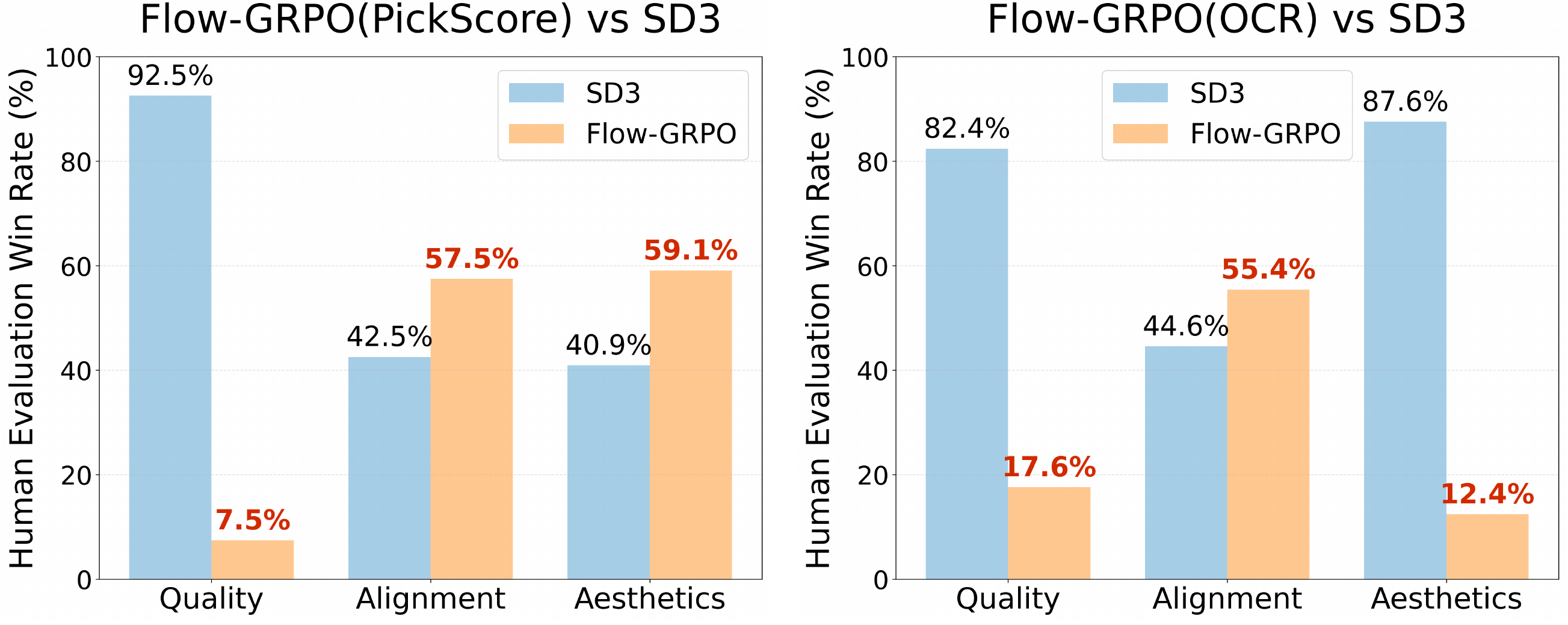}
    \caption{
        Human evaluation comparing Flow-GRPO and SD3 under PickScore and OCR rewards.
    }
    \label{fig:reward_hack}
    \vspace{-10pt}    
\end{wrapfigure}

However, despite these encouraging results, applying GRPO to T2I generation still faces fundamental challenges. The main difficulty lies in the misalignment between reward models and true human aesthetic preferences. In practice, many reward models produce scalar outputs that introduce biases toward specific visual attributes, such as oversaturated colors in CLIP-based, PickScore~\cite{pickscore}, or HPS~\cite{hpsv2,hpsv3} models, or excessive text emphasis in OCR-based rewards. As a result, the generator may exploit these biases, achieving higher reward scores without genuine quality improvement, a phenomenon known as \textbf{reward hacking}. As shown in Fig.~\ref{fig:reward_hack}, Flow-GRPO underperforms the base model in several aspects, such as lower image quality with the PickScore reward and reduced aesthetics and quality under the OCR reward in human evaluation.

A common remedy is to add Kullback-Leibler (KL) regularization to contrain the parameter updates, which reduces reward hacking but limits optimization and lowers performance. To address this, we introduce \textbf{Adv-GRPO}, a novel RL framework with an adversarial reward that iteratively updates both the reward model and the base model. We observe that many high-quality reference images receive low scores from existing reward models. Therefore, we incorporate reference images as high-quality supervision, training the reward model as a discriminator to distinguish them from generated samples. Meanwhile, the base model as a generator is optimized with the GRPO loss. For the reward, we focus on human-preference reward models (e.g., HPS~\cite{hpsv2,hpsv3}, PickScore~\cite{pickscore}, Aesthetic models~\cite{aestheticv25}), the main paradigm in T2I generation using our adversarial optimization. For other reward models, such as rule-based rewards (e.g., OCR), we leverage reference images in a multi-reward optimization scheme to enhance robustness. Our method consistently improves performance across both types, showing strong adaptability and generalization.

Although our method achieves better visual results and effectively mitigates reward hacking in existing reward models, some bias remains. For example, the PickScore reward tends to sacrifice image quality, while the OCR reward may reduce aesthetic fidelity. To address this, we directly use reference images as rewards by introducing a \textbf{reward model derived from a visual foundation model}~\cite{dinov2,clip,sa1b,sam2} to further optimize the T2I model. Specifically, we leverage the DINO~\cite{dinov2} to provide stronger visual signals. Within our adversarial optimization framework, DINO is fine-tuned as a reward model to guide the generator toward better visual alignment using reference samples. The output feature of DINO serves as the reward to optimize the base model. As a result, these dense visual signals, rather than a single scalar from the existing reward models, enable the base model to produce images with improved aesthetics, text alignment, and overall visual fidelity.

Moreover, we further introduce a new RL-based application for style transfer, where different reference datasets of distinct styles are used to effectively guide the base model toward specific visual domains. Finally, we conduct experiments across multiple benchmarks, and the results show that our method consistently improves image quality, text alignment, and aesthetics while maintaining comparable benchmark reward scores. Under the PickScore and OCR rewards, our method achieves winning rates of 70.0\% and 85.3\% in image aesthetics compared with Flow-GRPO in human evaluation. Under the DINO reward, it further achieves a 72.4\% winning rate in aesthetics compared with the base model in human evaluation.

Our main contributions are summarized as follows:
\begin{itemize}
\item We are the first to introduce an RL framework with an adversarial reward that leverages high-quality reference images to jointly optimize the T2I model and reward model.

\item We extend our approach to multiple types of existing reward models. Furthermore, we explore using visual foundation models as reward to guide the optimization of the base T2I model.


\item Extensive experiments show that our method effectively alleviates reward hacking in existing reward models while maintaining competitive performance on standard benchmarks. With visual foundation model rewards, our method achieves comprehensive improvements in image quality, text–image alignment, and aesthetics.

\end{itemize}

\section{Related Work}

\subsection{Reinforcement Learning for Image Generation}
Recently, online reinforcement learning (RL) has shown strong effectiveness in improving the capabilities of large language models (LLMs)~\cite{deepseek_r1,grpo,gspo} and multimodal LLMs (MLLMs)~\cite{grpo_vl,unirl,minio3,deepeyes,dr_grpo,pixel_reasoner}, and it has also been applied to text-to-image (T2I) generation. Compared with earlier RL methods such as PPO, GRPO is more efficient since it removes the need for an additional value network. In the context of T2I generation, prior methods such as DPO~\cite{dpo_diffusion, raft, video_dpo,human_feedback_finetune_diffusion, self_play_finetune, step_dpo} and PPO~\cite{dpo,diffusion_ppo,diffusion_ppo2, diffusion_ppo3, diffusion_ppo4} have demonstrated effectiveness, and GRPO has also been adapted in this domain. For instance, DanceGRPO~\cite{dancegrpo} applies GRPO to both image and video generation models~\cite{sd3,flux,hunyuan,wan,sdxl}, while Flow-GRPO~\cite{flowgrpo} modifies the optimization process by replacing the ODE~\cite{flowmatching_ode} with an SDE to improve sampling diversity. Despite these advances, such methods still face challenges including reward hacking and training instability. Building on Flow-GRPO, several work~\cite{flowcps,mixgrpo} further refines the SDE process to stabilize optimization. Prior studies have sought to improve reward reliability in RL-based image generation. Works such as~\cite{ICTHP,mps,hpsv3} reduce aesthetic bias through refined reward design, while SRPO~\cite{srpo} enhances efficiency using semantic positive–negative prompts. In contrast, we propose an adversarial training framework with reference samples.

\subsection{Reward Models for Image Quality Assessment}


In T2I generation, the main reward models are human-preference models, such as HPS~\cite{hpsv2,hpsv3}, HPDv2~\cite{hpdv2}, PickScore~\cite{pickscore}, and Aesthetic models~\cite{aestheticv25,laionAesthetics}, which are built upon CLIP~\cite{clip} and fine-tuned on large-scale human preference datasets. Other variants, like ImageReward~\cite{imagereward} and UnifiedReward~\cite{unifiedreward}, further refine aesthetic alignment.
In addition, rule-based rewards, such as OCR-based text accuracy and GenEval~\cite{geneval} for object correctness, provide explicit but narrow supervision.
However, both reward types are prone to reward hacking, as they often overfit to specific biases rather than capturing true perceptual quality.

\section{Method}

In this section, we first introduce the preliminaries of GRPO for flow matching and adversarial training in Sec.~\ref{sec:preliminary}. We then describe our proposed adversarial optimization framework in Sec.~\ref{sec:adversarial_training}. Finally, in Sec.~\ref{sec:dino_as_reward}, we extend our approach by leveraging a visual foundation model as the reward to further enhance overall image quality. An overview of the entire pipeline is illustrated in Fig.~\ref{fig:pipeline}.

\subsection{Preliminary}
\label{sec:preliminary}

\textbf{GRPO on Flow Matching.} The denoising process in diffusion models~\cite{diffusion,flow_matching,rectified_flow} can be viewed as a Markov Decision Process (MDP), 
where each reverse step from $x_t$ to $x_{t-1}$ is treated as an action sampled from a policy 
$\pi_\theta(\cdot|x_t, c)$ conditioned on the current noisy sample $x_t$ and the text prompt $c$. 
In practice, $\pi_\theta$ corresponds to the conditional flow distribution $p_\theta(x_{t-1}|x_t, c)$. 
At each iteration, we generate a group of $G$ samples $\{x_0^i\}_{i=1}^G$ from the previous policy 
$\pi_{\theta_{\text{old}}}$ and compute their rewards $R(x_0^i, c)$.
The group advantage $\hat{A}^i$ is obtained by normalizing the reward of each sample within the group:
\begin{equation}
\hat{A}^i = \frac{R(x_0^i, c) - \mathrm{mean}\big(\{R(x_0^j, c)\}_{j=1}^G\big)}{\mathrm{std}\big(\{R(x_0^j, c)\}_{j=1}^G\big)}.
\label{eq:advantage}
\end{equation}

GRPO then optimizes the policy model by maximizing the following objective:
\begin{equation}
f(r, \hat{A}, \theta, \epsilon, \beta)
=
\frac{1}{G} \sum_{i=1}^G 
\frac{1}{T} \sum_{t=0}^{T-1}
\min\!\big(r_t^i(\theta)\hat{A}^i,\; \tilde{r}_t^i(\theta)\hat{A}^i\big)
-\beta\, D_{\mathrm{KL}}\!\left(
\pi_\theta(\cdot|x_t^i, c)\,\big\|\,\pi_{\theta_{\text{old}}}(\cdot|x_t^i, c)
\right),
\label{eq:grpo-clip}
\end{equation}
with the importance ratio
\begin{equation}
r_t^i(\theta)
= 
\frac{
p_\theta(x_{t-1}^i \mid x_t^i, c)
}{
p_{\theta_{\text{old}}}(x_{t-1}^i \mid x_t^i, c)
},
\quad
\tilde{r}_t^i(\theta)
= 
\mathrm{clip}\!\big(
r_t^i(\theta),
1-\epsilon,
1+\epsilon
\big).
\label{eq:importance-ratio}
\end{equation}
Here, $\epsilon$ controls the clipping range and $\beta$ weights the KL penalty to stabilize training.

\noindent\textbf{Adversarial Training.}  
\label{sec:adversarial_training}
Adversarial training is typically formulated as a minimax optimization problem between a generator $G_\theta$ and a discriminator $D_\phi$:  
\begin{equation}
\min_{\theta} \max_{\phi} \; 
\mathbb{E}_{x \sim p_{\text{data}}} \big[ \log D_\phi(x) \big] 
+ \mathbb{E}_{z \sim p_z} \big[ \log \big(1 - D_\phi(G_\theta(z)) \big) \big],
\label{eq:adv-general}
\end{equation}
where $p_{\text{data}}$ denotes the real data distribution and $p_z$ is a prior distribution over latent variables. The discriminator $D_\phi$ is trained to distinguish real samples from generated ones, while the generator $G_\theta$ is optimized to produce samples that can fool the discriminator.


%



\subsection{GRPO with Adversarial Reward}

\begin{wrapfigure}{r}{0.55\linewidth} 
    \vspace{-10pt}  
    \centering
    \includegraphics[width=\linewidth]{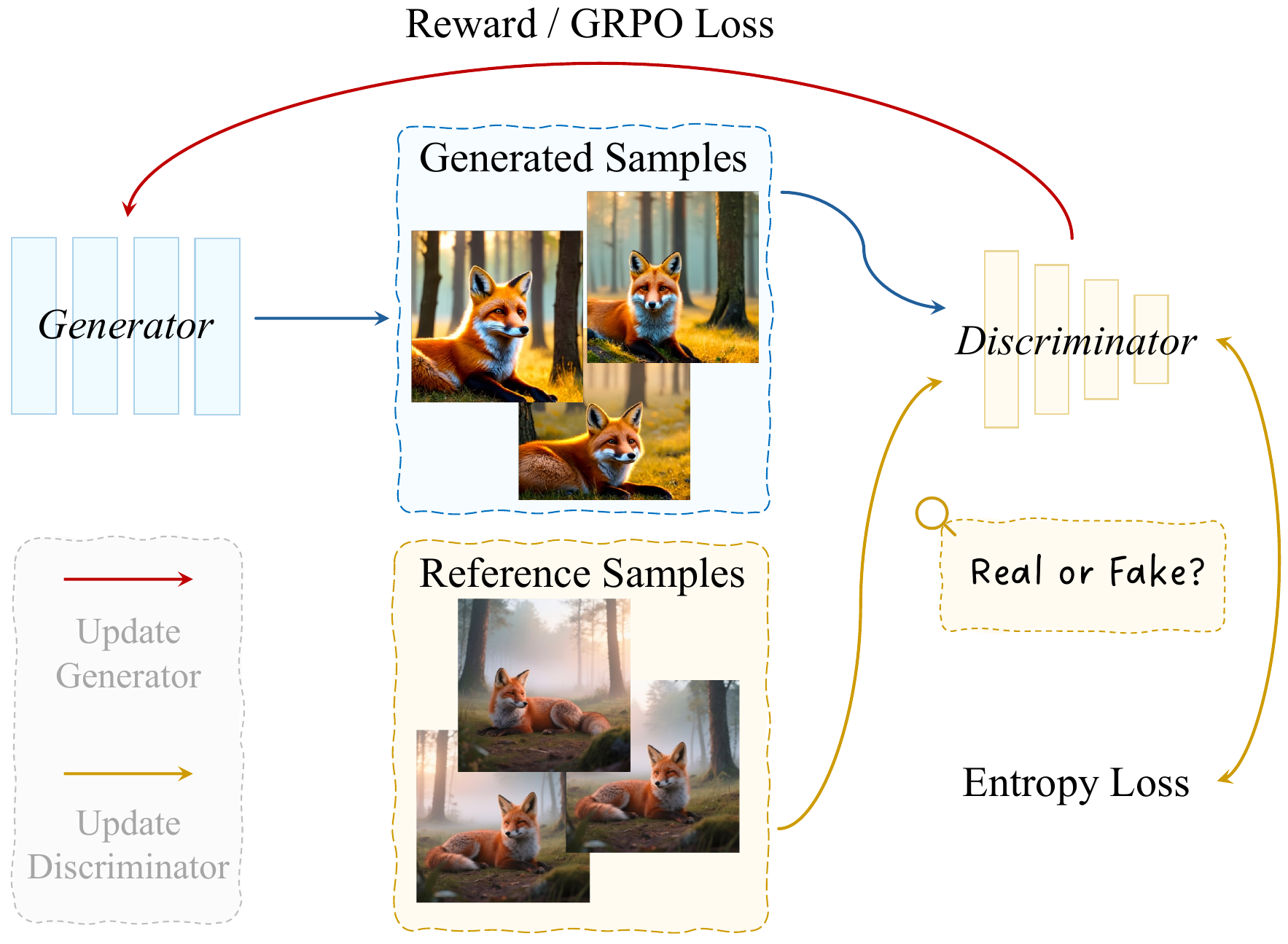}
    \vspace{-12pt}
    \caption{
        Pipeline of \method{}. 
        The generator is optimized using the GRPO loss, while the discriminator is trained to distinguish between generated samples and reference images, treated as negative and positive samples, respectively.
        The discriminator serves as a reward model to provide feedback for the generator.
    }
    \label{fig:pipeline}
    \vspace{-10pt} 
\end{wrapfigure}

As shown in Fig.~\ref{fig:pipeline}, our method extends GRPO to an adversarial setting, where the text-to-image (T2I) generator and the reward model are jointly optimized.
The generator $G_\theta$ is trained via GRPO to maximize the rewards of its generated samples. The reward model $R_\phi$ serves as a discriminator, adversarially trained to distinguish high-quality reference images from generated ones.
Specifically, given a text prompt $c$, the generator produces a group of samples 
$\{x_g^i = G_\theta(c)\}_{i=1}^G$ with corresponding reward values $R_\phi(x_g^i, c)$. 
The generator is optimized under the standard GRPO objective:
\begin{equation}
J_{\text{gen}}(\theta) 
= \mathbb{E}_{c \sim \mathcal{C}, \{x_g^i\}_{i=1}^G \sim G_{\theta_{\text{old}}}} 
\Big[ f(r, \hat{A}, \theta, \epsilon, \beta) \Big],
\label{eq:adv-grpo-gen}
\end{equation}
where $f(r, \hat{A}, \theta, \epsilon, \beta)$ follows the clipped GRPO formulation described in Eq.~\ref{eq:grpo-clip}, 
and $\hat{A}$ denotes the normalized group advantage.  

Meanwhile, the reward model is optimized using reference high-quality data $\mathcal{D}_{\text{ref}} = \{x_r\}$ as positive samples 
and generated images $\{x_g\}$ as negative samples:
\begin{equation}
J_{\text{reward}}(\phi)
= -\mathbb{E}_{x_r \sim \mathcal{D}_{\text{ref}}}\!\left[\log R_\phi(x_r)\right]
 - \mathbb{E}_{x_g \sim G_\theta(c)}\!\left[\log(1 - R_\phi(x_g))\right].
\label{eq:adv-grpo-reward}
\end{equation}
This adversarial co-training enforces the reward model to align with reference image distributions while guiding the generator toward higher-quality outputs.

This joint training objective establishes a dynamic equilibrium: the generator strives to produce images that maximize the reward, while the reward model continuously adapts by contrasting generated samples with high-quality references. In this process, reward hacking is effectively mitigated, as $R_\phi$ learns to better reflect perceptual quality beyond its initial biases, and $G_\theta$ is encouraged to improve both reward scores and overall visual fidelity.

\noindent\textbf{Human Preference Models.} Human preference models are the primary type of reward function used in current T2I generation.
They are trained on human-labeled data, where annotators provide pairwise comparisons or aesthetic judgments to capture subjective visual preferences. We adopt such an adversarial co-training mechanism that leverages reference high-quality samples. 
Let $R(\cdot)$ denote the reward model, and let $G$ denote the generator. 
We monitor the average reward scores for generated and reference samples respectively:
\begin{equation}
\bar{r}_{\text{gen}} = \mathbb{E}_{x_g \sim G}[R(x_g)], \quad 
\bar{r}_{\text{ref}} = \mathbb{E}_{x_r \sim D_{\text{ref}}}[R(x_r)].
\end{equation}

When the average reward of generated images surpasses that of reference images, i.e., $\bar{r}_{\text{gen}} > \bar{r}_{\text{ref}}$, we regard this as a signal of potential reward hacking. At this point, we trigger adversarial fine-tuning of the reward model, where reference samples are treated as positive samples and generated samples as negative samples. This process re-aligns the reward model toward human-preferred visual quality and prevents degenerate feedback loops during GRPO optimization. The optimization objective is defined in Eq.~\ref{eq:adv-grpo-reward}.


\noindent\textbf{Rule-based Reward Models.}
Besides the main human-preference reward models, other rewards such as rule-based metrics (e.g., OCR or GenEval~\citep{geneval}) provide clear task-specific signals but are inherently deterministic and non-differentiable, making them unsuitable for adversarial training. To address this, we fully leverage high-quality reference images and adopt a simple multi-reward formulation to balance task specificity and visual realism. The reward is 
\begin{equation}
R_{\text{combined}}(x_g, c) 
= \lambda \, R_{\text{rule}}(x_g, c) 
+ (1-\lambda) \, \mathrm{sim}_{\text{CLIP}}(x_g, x_r),
\label{eq:multi-reward}
\end{equation}
where $R_{\text{rule}}$ denotes the task-specific reward (e.g., OCR or GenEval), 
$\mathrm{sim}_{\text{CLIP}}$ measures the CLIP similarity between the generated image $x_g$ and a reference image $x_r$, 
and $\lambda \in [0,1]$ controls the trade-off. This formulation stabilizes training by preventing rule-based objectives from dominating and preserving overall visual fidelity.

\subsection{Visual Foundation Models As Reward}
\label{sec:dino_as_reward}
The existing reward models provide explicit supervision but capture only limited aspects of image quality and often introduce aesthetic or content biases. Incorporating reference images via adversarial co-training alleviates reward hacking but mainly regularizes the reward model rather than holistically improving image quality. 

Therefore, we further explore using \textbf{visual foundation models} as reward models to guide the optimization of the base generator. Unlike conventional reward models, visual foundation models encode rich semantic and structural representations of natural images, making them well-suited for aligning the overall distribution of generated images with that of high-quality reference images.





Formally, given a pre-trained visual foundation model $F_\psi(\cdot)$ (e.g., DINO~\citep{dinov2}), 
we freeze its parameters and attach a lightweight binary classification head $h_\phi(\cdot)$ on top of its representations. 
For each input image $x$, we extract both the global [CLS] embedding and the patch-level features:
\begin{equation}
\mathbf{f}_{\text{cls}}, \mathbf{F}_{\text{patch}} = F_\psi(x),
\label{eq:dino_features}
\end{equation}
where $\mathbf{f}_{\text{cls}} \in \mathbb{R}^{D}$ denotes the [CLS] token feature, and 
$\mathbf{F}_{\text{patch}} \in \mathbb{R}^{N \times D}$ represents the $N$ patch embeddings. 


Given the global [CLS] feature $\mathbf{f}_{\text{cls}}$ and patch-level features 
$\mathbf{F}_{\text{patch}} = \{\mathbf{f}_j\}_{j=1}^{N}$ extracted from the frozen visual backbone $F_\psi(\cdot)$,  
the reward is computed using a shared classification head $h_\phi(\cdot)$ as:
\begin{equation}
R_{\text{global}}(x) = h_\phi\!\left(\mathbf{f}_{\text{cls}}\right), \quad
R_{\text{local}}(x) = \frac{1}{n} \sum_{j \in \mathcal{S}} h_\phi\!\left(\mathbf{f}_j\right),
\label{eq:reward_global_local_single_head}
\end{equation}
where $\mathcal{S} \subset \{1, \ldots, N\}$ denotes a randomly selected subset of $n$ patch tokens.  
This stochastic sampling encourages the model to focus on diverse local structures while maintaining computational efficiency. The final reward combines both components:
\begin{equation}
R_\phi(x) = \lambda_{\text{g}} R_{\text{global}}(x) 
+ \lambda_{\text{l}} R_{\text{local}}(x),
\label{eq:reward_final}
\end{equation}
where $\lambda_{\text{g}}$ and $\lambda_{\text{l}}$ control the relative contribution of global and local cues.



During adversarial training, the reward head $h_\phi$ is trained to discriminate reference images $x_r \sim \mathcal{D}_{\text{ref}}$ (positives) from generated images $x_g \sim G_\theta(c)$ (negatives). We employ a hinge loss objective for this discrimination. Specifically, $h_\phi$ is applied to both the global [CLS] feature and a subset of randomly sampled patch features extracted from the frozen backbone $F_\psi(\cdot)$, with separate hinge losses computed at the global and local levels. The corresponding hinge losses at the global and local levels are defined as:
\begin{equation}
\mathcal{L}_{\text{global}}(\phi)
= \mathbb{E}_{x_r}\!\left[\max(0,\,1-h_\phi(\mathbf{f}_{\text{cls}}^r))\right]
+ \mathbb{E}_{x_g}\!\left[\max(0,\,1+h_\phi(\mathbf{f}_{\text{cls}}^g))\right];
\end{equation}

\begin{equation}
\mathcal{L}_{\text{local}}(\phi)
= \mathbb{E}_{x_r}\!\left[\tfrac{1}{|\mathcal{S}|}\!\sum_{j\in\mathcal{S}}\max(0,\,1-h_\phi(\mathbf{f}_j^r))\right]
+ \mathbb{E}_{x_g}\!\left[\tfrac{1}{|\mathcal{S}|}\!\sum_{j\in\mathcal{S}}\max(0,\,1+h_\phi(\mathbf{f}_j^g))\right].
\label{eq:hinge_losses}
\end{equation}
The final adversarial loss for training the reward model is a weighted combination:
\begin{equation}
\mathcal{L}_{\text{reward}}(\phi) 
= \lambda_{\text{g}} \mathcal{L}_{\text{global}}(\phi) 
+ \lambda_{\text{l}} \mathcal{L}_{\text{local}}(\phi),
\label{eq:final_reward_loss}
\end{equation}
where $\mathbf{f}_{\text{cls}}^{r}$ and $\mathbf{f}_{\text{cls}}^{g}$ denote 
the global features of reference and generated images, 
and $\mathbf{f}_{j}^{r}$, $\mathbf{f}_{j}^{g}$ represent their patch-level features. 

This global–local reward formulation enables the generator to benefit from both complementary aspects: the global [CLS] feature emphasizes high-level semantics and structural consistency, while the local patch features capture fine-grained texture and detail. Together, they allow the model to generate more coherent and visually refined images.

\section{Experiments}

\begin{figure*}[t]
    \centering
    \includegraphics[width=1\textwidth]{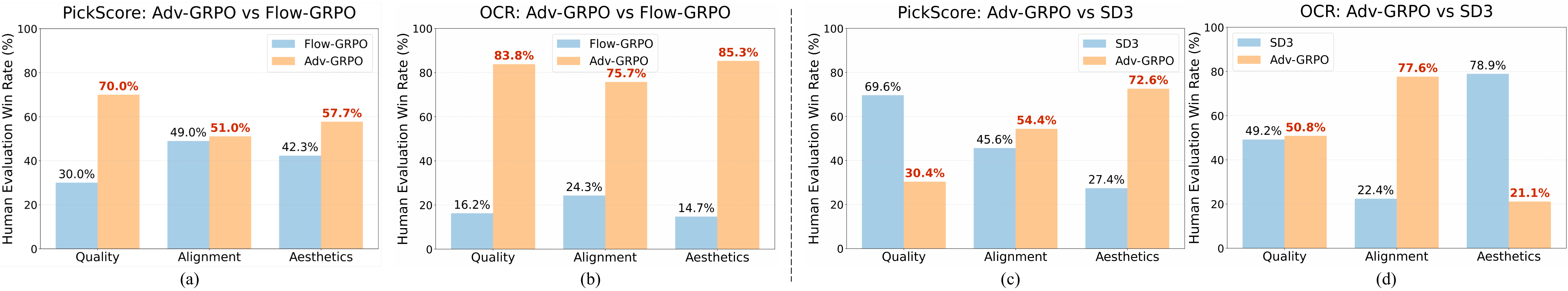}
    \vspace{-20pt}
    \caption{
        Human evaluation under PickScore- and OCR-based rewards. Our method \method{} improves image quality and aesthetics with PickScore reward in \textbf{a)}, and for all metrics with OCR reward in \textbf{b)}.
        Compared with the original model (SD3), PickScore reward trade-off aesthetic improvements with image quality degradation in \textbf{c)}, OCR reward trade-off text-alignment from aesthetics degradation in \textbf{d)}.
        }
    \vspace{-10pt}
    \label{fig:human_eval}
\end{figure*}

\begin{figure*}[t]
    \centering
    \includegraphics[width=1\textwidth]{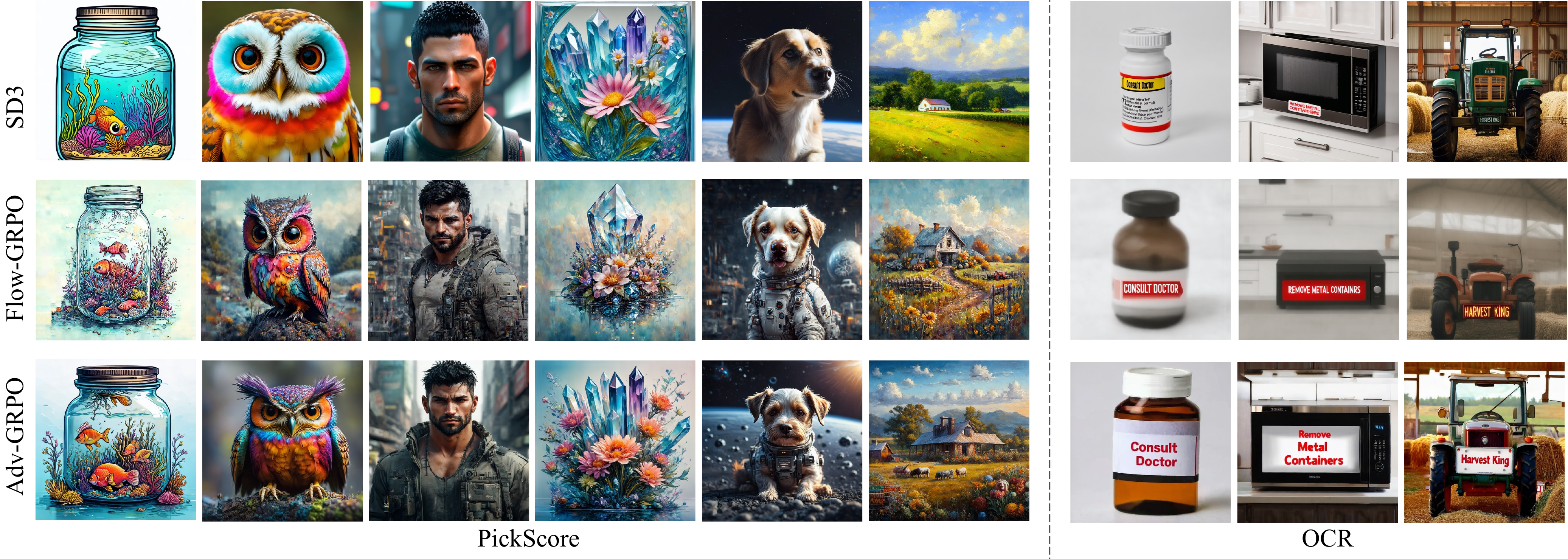}
    \vspace{-20pt}
    \caption{
        Visualizations under PickScore (\textbf{Left}) and OCR (\textbf{Right}) rewards. Our method \method{} alleviates reward hacking for both. 
    }
    \vspace{-15pt}
    \label{fig:visualization_pickscore_ocr}
\end{figure*}

\subsection{Implementation Details}
\noindent\textbf{Training Setup.} We adopt Stable Diffusion 3 (SD3)~\cite{sd3} as the base generator.
For the PickScore~\cite{pickscore} reward, we use the PickScore prompt dataset for training and evaluation,
and for the OCR reward, we employ the OCR prompt set. For visual foundation model experiments, we employ DINOv2~\cite{dinov2} as the reward model to optimize the base generator.
Under the DINO~\cite{dinov2} reward, our method is validated on PickScore, OCR, and GenEval prompts.
Each prompt forms a group of 16 samples during training.
We fine-tune only the last two layers of PickScore’s vision branch (learning rate $3\times10^{-4}$ for the generator and $5\times10^{-6}$ for the reward model) for 1{,}000 iterations.
For DINO, we train the classification head with a learning rate of $1\times10^{-4}$.
In the OCR setting, we jointly optimize SD3 using both OCR and CLIP similarity rewards.
Training uses 10 inference steps, with 2 timesteps randomly sampled from the 50–100\% noise schedule.
Eight reference images per prompt are generated with Qwen-Image~\cite{qwen_image}. All experiments are conducted on 8 NVIDIA H100 GPUs. Further details are provided in the supplementary material.

\noindent\textbf{Baselines.}
We compare our method with two baselines: 
\textit{Base Model}, the original SD3 without reinforcement learning optimization; and \textit{Flow-GRPO}~\cite{flowgrpo}, a GRPO-based variant of SD3 that reformulates diffusion sampling as a stochastic differential equation to enhance training stability and diversity.

\subsection{Evaluation Protocol}

\noindent\textbf{Metrics.}
We evaluate our method using these reward metrics: PickScore, OCR accuracy, and GenEval score~\cite{geneval}. In addition, we also compute the DINO similarity, which measures the cosine similarity between image embeddings extracted by the DINO, reflecting the semantic consistency between generated and reference images.

\noindent\textbf{Human Evaluation.}
In addition, we conduct a comprehensive human evaluation covering three aspects: \textit{Aesthetics}, \textit{Alignment}, and \textit{Quality}. The aesthetic score measures overall visual appeal and artistic composition. The alignment score evaluates the semantic consistency between generated images and text prompts, while the quality score reflects perceptual fidelity, structural coherence, and the absence of artifacts or distortions. We employ 12 expert evaluators to perform pairwise comparisons across 100 prompts for each reward setting, covering a total of 400 diverse prompts. In total, 10 comparison pairs across different rewards and methods are evaluated, resulting in 12{,}000 human comparison judgments. To ensure evaluation reliability, we conduct expert calibration, resolve inconsistent annotations, and continuously verify scoring criteria during the evaluation process. Further details of the evaluation protocol are provided in the supplementary material.

\subsection{Main Results}

\begin{figure*}[t]
    \centering
    \includegraphics[width=1\textwidth]{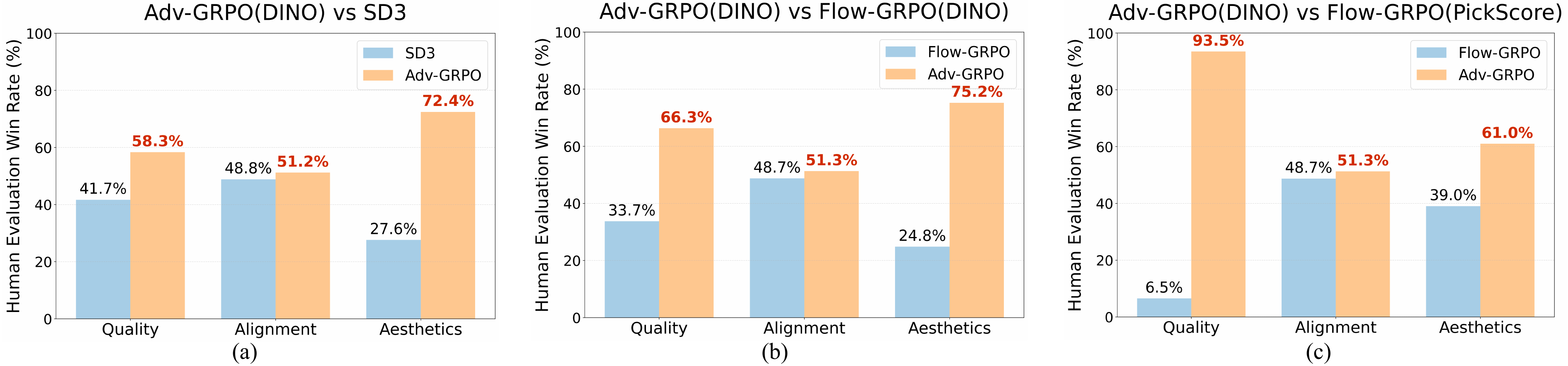}
    \vspace{-15pt}
    \caption{Human evaluation results under the visual foundation model (DINO) reward.
    Using a foundation model as the reward, our RL method improves image aesthetics, quality, and text alignment compared with the original SD3 model (a), and significantly outperforms Flow-GRPO under the DINO similarity reward (b) and PickScore reward (c).}
    \vspace{-10pt}
    \label{fig:human_eval_dino}
\end{figure*}

\begin{figure*}[t]
    \centering
    \includegraphics[width=1\linewidth]{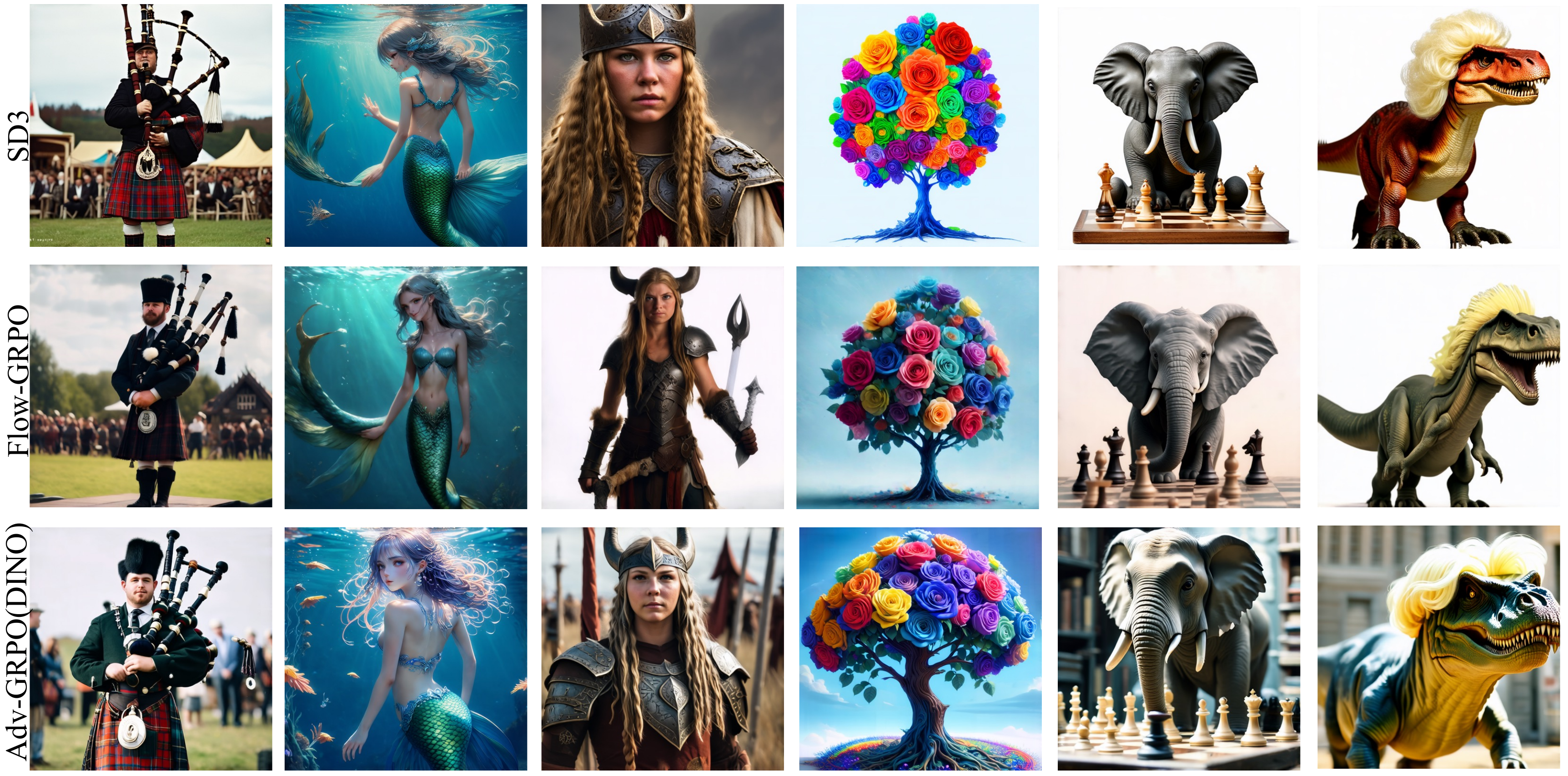}
    \caption{
        Visualizations under the DINO reward model.
        With \textbf{adversarial DINO reward}, our method shows better visual quality.
    }
    \vspace{-10pt}
    \label{fig:visualization_dino}
\end{figure*}

\noindent\textbf{Reward Hacking Mitigation.}


\begin{wraptable}{r}{0.55\linewidth}  
    \vspace{-15pt}  
    \centering
    \footnotesize
    \caption{
        Comparison under different reward models.
        Each row corresponds to an independent optimization using the specific reward and its associated evaluation metric.
    }
    \vspace{-4pt}
    \setlength{\tabcolsep}{6pt}
    \resizebox{\linewidth}{!}{
    \begin{tabular}{llccc}
        \toprule
        \multirow{2}{*}{Reward Model} & \multirow{2}{*}{Metric} & \multicolumn{3}{c}{Method} \\
        \cmidrule(lr){3-5}
         &  & SD3 & Flow-GRPO & \method{} \\
        \midrule
        PickScore & PickScore $\uparrow$ & 21.70 & 22.82 & 22.78 \\
        OCR       & OCR Accuracy $\uparrow$ & 0.58  & 0.91  & 0.91 \\
        \bottomrule
    \end{tabular}
    }
    \label{tab:reward_specific}
    \vspace{-10pt} 
\end{wraptable}

We evaluate reward hacking from two perspectives. 
\textit{a) Comparable benchmark performance with Flow-GRPO.}
As shown in Tab.~\ref{tab:reward_specific}, our method achieves comparable benchmark scores to Flow-GRPO, indicating that adversarial training does not compromise quantitative performance in corresponding metrics.
For PickScore, both methods reach around 22.80, substantially higher than 21.70 from SD3.
For OCR, our method and Flow-GRPO achieve 0.91 accuracy, outperforming SD3 (0.58) by a large margin.
This demonstrates that our approach maintains strong quantitative results while addressing reward bias.

\noindent\textit{b) Significant improvement on human evaluation and visualizations.}
As shown in Fig.~\ref{fig:human_eval}(a)(b), our method consistently outperforms Flow-GRPO under both PickScore and OCR rewards, achieving higher aesthetic, alignment, and quality scores.
In particular, the win rate reaches 70\% in image quality under PickScore and 85.3\% in aesthetics under OCR. Compared with SD3 (Fig.~\ref{fig:human_eval}(c)(d)), our method achieves a 72.6\% win rate in aesthetics under the PickScore reward and a 77.6\% win rate in alignment under the OCR reward, demonstrating substantial perceptual improvements. However, we also observe that PickScore optimization tends to sacrifice image quality, while OCR optimization slightly compromises aesthetics, indicating that some inherent bias in these reward models remains. Visualizations in Fig.~\ref{fig:visualization_pickscore_ocr} further confirm that our approach produces images with better perceptual quality and overall fidelity.

\begin{wrapfigure}{r}{0.55\linewidth} 
    \vspace{-10pt}
    \centering
    \includegraphics[width=\linewidth]{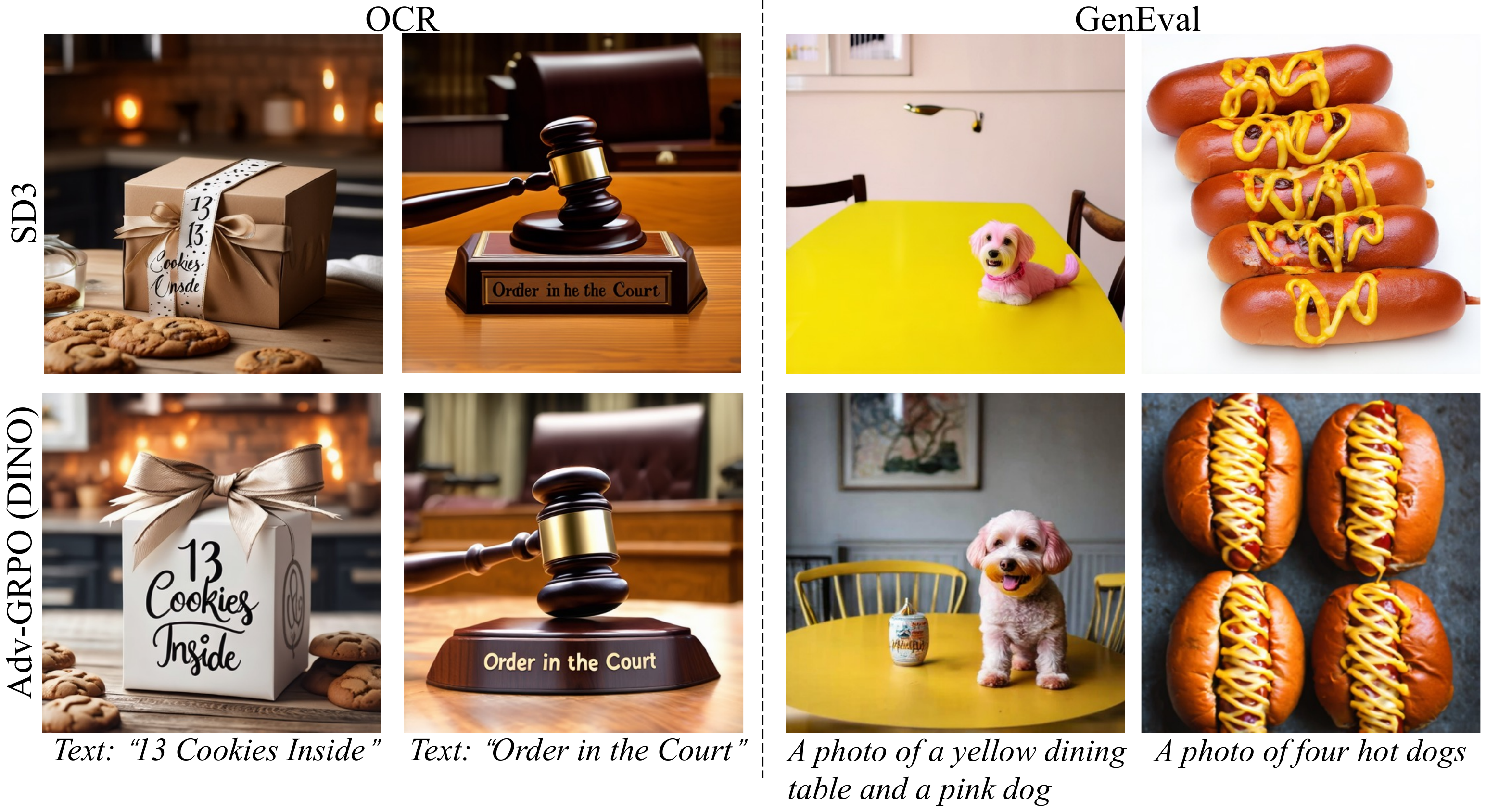}
    \vspace{-8pt}
    \caption{
        Visual comparison between our method (DINO reward)
        and SD3 across different task prompts.
    }
    \label{fig:dino_ocr_geneval}
    \vspace{-10pt}
\end{wrapfigure}

\noindent\textbf{Vision Foundation Model as Better Rewards.}
We evaluate using DINO as reward models, and compare our method with Flow-GRPO (DINO similarity as the reward) and SD3.
\textit{a) Comprehensive improvements without degration.}
Compared with SD3 in Fig.~\ref{fig:human_eval_dino}(a), our method consistently improves all human evaluation metrics, including aesthetics, alignment, and quality, especially the aesthetic dimension with 72.4\% win rate. 
Compared with Flow-GRPO (using DINO similarity as the reward) in Fig.~\ref{fig:human_eval_dino}(b), our method achieves a 66.3\% win rate in quality and a 75.2\% win rate in aesthetics.
Fig.~\ref{fig:human_eval_dino}(c) compares our method with Flow-GRPO (using PickScore reward), and  our method achieves 93.5\% win rate in quality. These results suggest that, compared with preference-based reward models,  using a visual foundation model (DINO etc.) as the reward provides a more comprehensive and reliable guidance  for image generation. The visualization results in Fig.~\ref{fig:visualization_dino} also show that our approach produces higher-quality images with richer backgrounds and improved aesthetics.


\begin{wraptable}{r}{0.52\linewidth} 
    \vspace{-10pt}
    \centering
    \small
    \caption{
        General evaluation using the DINO reward across multiple tasks,
        comparing our method with SD3.
    }
    \vspace{-4pt}
    \setlength{\tabcolsep}{6pt}
    \resizebox{\linewidth}{!}{
    \begin{tabular}{lccc}
        \toprule
        Method & PickScore $\uparrow$ & OCR Accuracy $\uparrow$ & GenEval $\uparrow$ \\
        \midrule
        SD3 & 21.70 & 0.59 & 0.61 \\
        \method{} (DINO) & 21.90 & 0.69 & 0.69 \\
        \bottomrule
    \end{tabular}
    }
    \label{tab:dino_compare_sd3}
    \vspace{-10pt}
\end{wraptable}

\noindent\textit{b) Consistent improvement across benchmarks.}
We validate the versatility of the DINO reward using different benchmark prompts, including OCR and GenEval.
As shown in Tab.~\ref{tab:dino_compare_sd3}, our adversarial DINO reward consistently improves performance across tasks, increasing OCR accuracy from 0.59 to 0.69 and GenEval score from 0.61 to 0.69 compared with SD3.
The visual results in Fig.~\ref{fig:dino_ocr_geneval} also demonstrate visually appealing outputs, confirming that DINO serves as a general and reliable reward model across diverse objectives.
 Our reward curves exhibit a steady increase over training iterations, converging within roughly 1,000 steps, which is provided in the supplimentary material.

\subsection{Ablation Study}

\begin{wrapfigure}{r}{0.3\linewidth} 
    \vspace{-10pt}
    \centering
    \includegraphics[width=\linewidth]{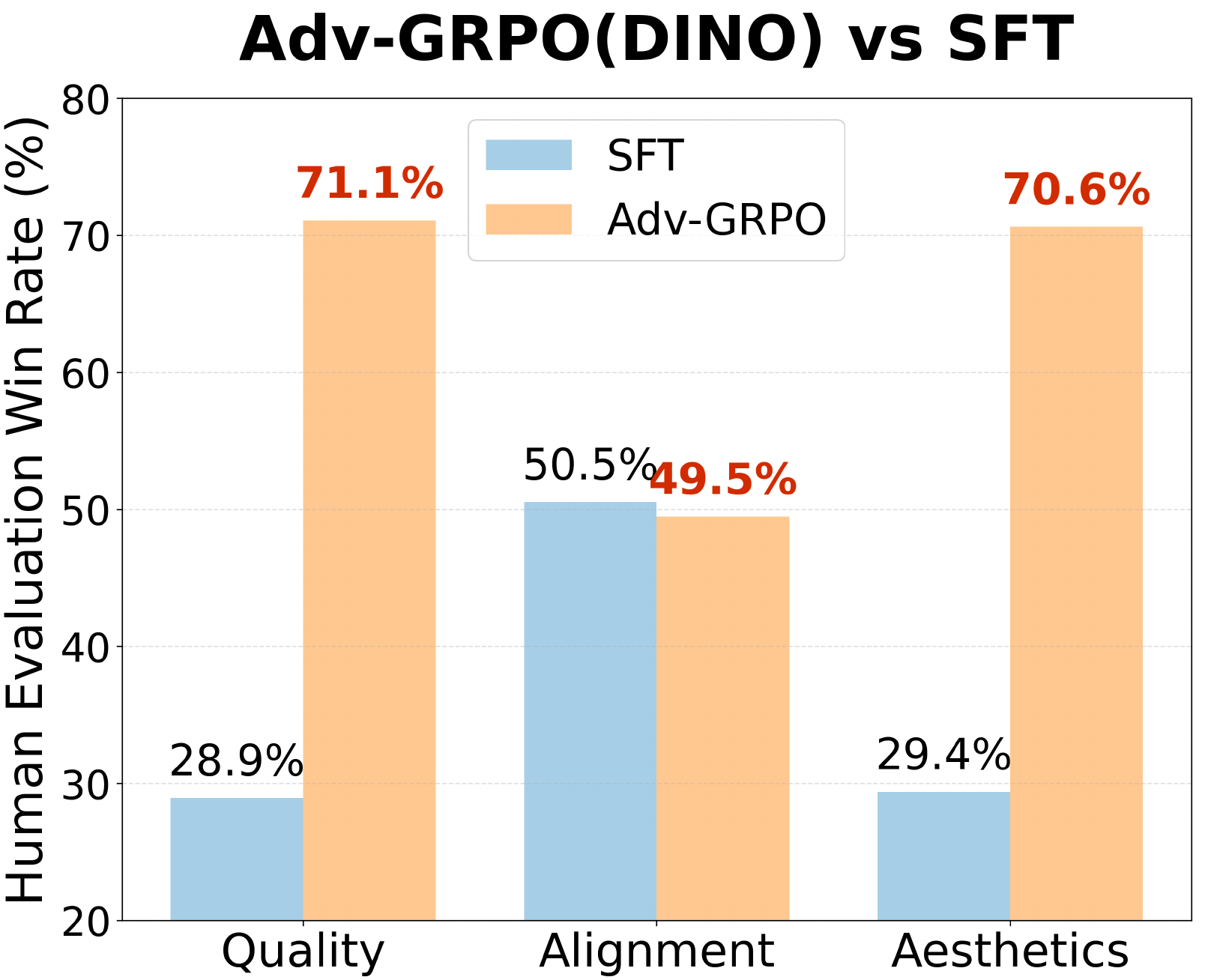}
    \vspace{-15pt}
    \caption{
         Human evaluation comparing our DINO-reward model with SFT, where our method performs better.
    }
    \label{fig:ablation_vis}
    \vspace{-15pt}
\end{wrapfigure}

\textbf{Number of Reference Images.}
We study the effect of the number of reference images by varying the dataset size across 200, 500, and 1,000 samples. As shown in Tab.~\ref{tab:ablation_refnum}, our method achieves comparable DINO similarity even with only 200 reference images, indicating that a small dataset is sufficient for effective optimization. The qualitative results in Fig.~\ref{fig:ablation_vis} further show that visual quality and style consistency remain stable as the number of references increases, demonstrating the data efficiency of our approach.




\begin{figure}[h!]
    \centering
    \includegraphics[width=1\linewidth]{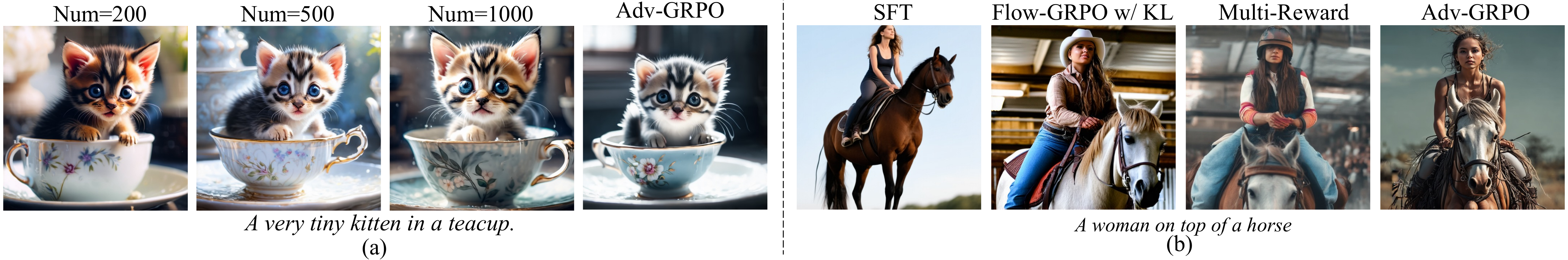}
    \caption{
        Ablation results. (a) Visualizations with different
    numbers of reference images, showing effectiveness even
    with 200 samples. (b) Visualizations of ablation studies on SFT, KL regularization, multi-reward optimization, and our method \method{}.
    }
    \vspace{-10pt}
    \label{fig:ablation_vis2}
\end{figure}

\noindent\textbf{Comparison with Supervised Fine-Tuning (SFT).}
Human evaluation shows that our method under DINO reward model achieves notably higher perceptual quality than SFT, with over 70\% win rates in both aesthetics and image quality in Fig.~\ref{fig:human_eval_dino}(d). As shown in Fig.~\ref{fig:ablation_vis2} and Tab.~\ref{tab:kl_ablation}, our approach also attains better visual results and higher quantitative metrics. Unlike SFT, which cannot explicitly optimize for specific reward objectives, our RL framework enables targeted optimization toward desired aspects such as text readability or visual appeal.

\begin{table}[h!]
    \centering
    \scriptsize   
    \caption{
        Ablation on the number of reference samples used during inference.
        Our method maintains stable DINO similarity even with few reference images,
        demonstrating strong data efficiency.
    }
    \setlength{\tabcolsep}{4pt}  
    \resizebox{0.6\linewidth}{!}{
    \begin{tabular}{lccccc}
        \toprule
        \multirow{2}{*}{Metric} & \multirow{2}{*}{SD3} & \multicolumn{3}{c}{w/ Fewer Reference Samples} & \multirow{2}{*}{\method{}} \\
        \cmidrule{3-5}
         &  & 200 & 500 & 1000 &  \\
        \midrule
        DINO Similarity $\uparrow$ & 0.592 & 0.621 & 0.618 & 0.619 & 0.621 \\
        \bottomrule
    \end{tabular}
    }
    \label{tab:ablation_refnum}
\end{table}

\begin{table}[h!]
    \centering
    \scriptsize
    \caption{
        Ablation on SFT, KL regularization, and multi-reward optimization
        under PickScore and OCR metrics.
    }
    \setlength{\tabcolsep}{4pt}
    \resizebox{0.6\linewidth}{!}{
    \begin{tabular}{lcccc}
        \toprule
        Metric & SFT & Flow-GRPO (w/ KL) & Multi-Reward & \method{} \\
        \midrule
        PickScore $\uparrow$ & 21.60 & 21.84 & 21.60 & \textbf{22.78} \\
        OCR Accuracy $\uparrow$ & 0.68 & 0.80 & 0.91 & \textbf{0.91} \\
        \bottomrule
    \end{tabular}
    }
    \label{tab:kl_ablation}
\end{table}

\noindent\textbf{KL Regularization.}
We compare our method with Flow-GRPO using a KL regularization term. 
As shown in Tab.~\ref{tab:kl_ablation} and Fig.~\ref{fig:ablation_vis2}, 
adding a KL constraint leads to lower reward scores and degraded visual quality. 
KL regularization is sensitive, an overly large coefficient restricts optimization, while a small one cannot prevent reward hacking. Overall, our method achieves better stability and visual fidelity without relying on such fragile regularization.

\noindent\textbf{Multi-Reward Combination.}
We also compare our method with Flow-GRPO trained using a combination of PickScore and OCR rewards.
Although multi-reward optimization can also reduce reward hacking, balancing different reward weights is challenging due to varying sensitivities across models.
As shown in Tab.~\ref{tab:kl_ablation} and Fig.~\ref{fig:ablation_vis2}, our method achieves higher metrics and better visual fidelity than the multi-reward baseline.

\begin{figure}[h!]
    \centering
    \includegraphics[width=0.55\linewidth]{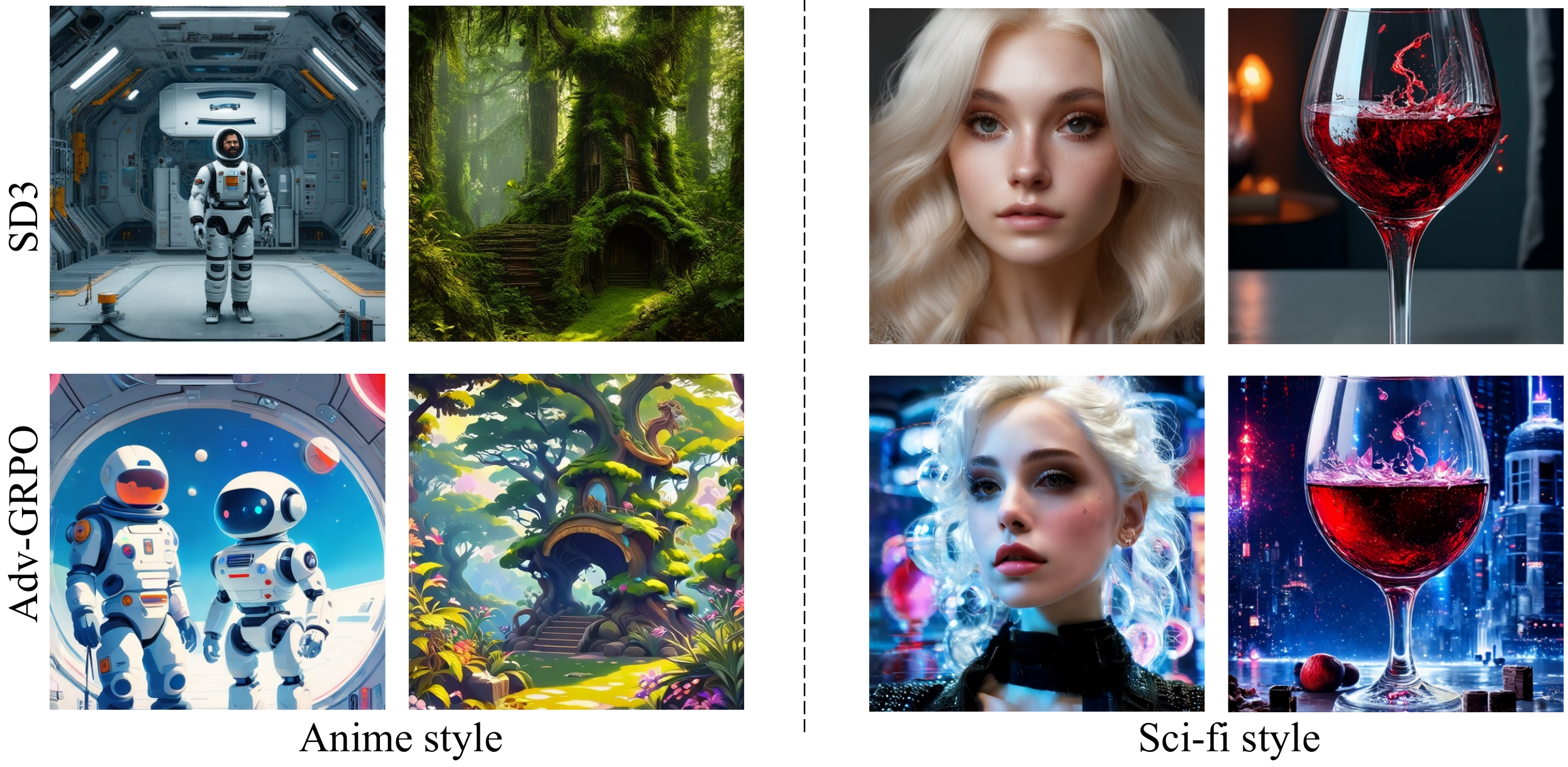}
    \caption{
        Application: Style transfer with the adversarial DINO reward.
        Our method successfully transfers the SD3 model to target visual domains,
        including \textbf{Anime} and \textbf{Sci-Fi} styles.
    }
    \label{fig:application}
\end{figure}

\subsection{Style Customization via Adv-GRPO}
\label{application}
We further demonstrate the versatility of our method through a style customization task. Unlike conventional RL-based T2I approaches that rely solely on text prompts and self-generated samples, our framework enables RL-driven optimization directly from pure image inputs, using visual foundation models such as DINO to guide learning. We construct two reference datasets, one anime-style and one sci-fi themed, and fine-tune the SD3 model with our proposed pipeline. As shown in Fig.~\ref{fig:application}, our method effectively transfers the generation style toward the reference domains while preserving semantic structure and image quality. This experiment showcases the flexibility and generalization of our approach, representing the first RL-based framework capable of performing style customization.



\section{Conclusion}
We introduce an RL framework with an adversarial reward for T2I generation. By leveraging reference high-quality references, the reward model better aligns with human visual preferences and mitigates reward hacking. Besides, incorporating visual foundation models such as DINO further provides unbiased visual guidance, improving overall image quality, aesthetics and text alignment. Extensive experiments verify the effectiveness and generality of our framework across diverse reward settings.

\section{Acknowledgement}

We thank Danze Chen and Kaiming Yang for their support in reference data generation. We are also grateful to Jiaming Han, Yuang Ai, and Shaobin Zhuang for their valuable advice and insightful discussions.



\bibliographystyle{plainnat}
\bibliography{main}

\clearpage

\beginappendix

In this supplementary material, we provide additional visualization results in Sec.~\ref{visualizations_our_method_more}, further style transfer examples in Sec.~\ref{style_transfer}, experiments using SigLIP for optimization in Sec.~\ref{siglip_optimization}, more implementation details in Sec.~\ref{more_implementation_details}, the reward curves in Sec.~\ref{reward_curve}, and the full procedures of our human evaluation in Sec.~\ref{more_human_evaltion}.

\section{More Implementation Details}
\label{more_implementation_details}

In the DINO reward, we assign a 7:3 weighting ratio to the global and local batch losses and rewards. For SD3, we apply LoRA-based fine-tuning with a configuration that uses a rank of 32, a scaling factor (lora\_alpha) of 64, and Gaussian initialization for all LoRA weights. During both training and evaluation, we set the classifier-free guidance (CFG) scale to 4.5, and employ bfloat16 mixed precision throughout the process.
For the DINO reward training schedule, we adopt a 10:1 update ratio, meaning that the discriminator is updated for 10 steps for every 1 generator step.
For the PickScore reward model, we perform fine-tuning only when the reward assigned to the generated images surpasses that of the reference images.

\section{Visualizations Under Our Method}
\label{visualizations_our_method_more}

\begin{figure*}[t!]
    \centering
    \includegraphics[width=1\textwidth]{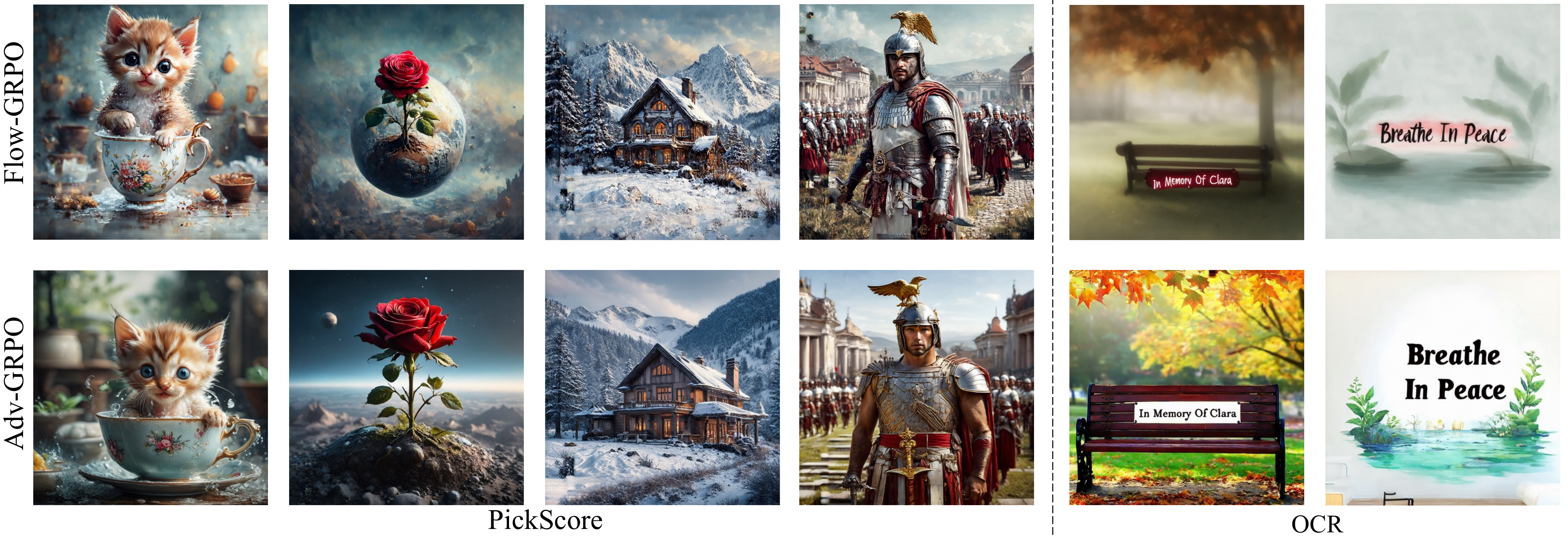}
    \vspace{-20pt}
    \caption{
        More Visualizations about alleviating reward hacking under PickScore and OCR reward models.
    }
    \vspace{-5pt}
    \label{fig:reward_hack_more}
\end{figure*}

\textbf{Alleviating Reward Hacking.} We provide additional visualizations to further demonstrate the effectiveness of our method across various reward models. As shown in Fig.~\ref{fig:reward_hack_more}, our approach significantly alleviates reward hacking issues present in existing reward models such as PickScore and OCR, producing images with consistently higher overall visual quality compared with Flow-GRPO.

\begin{figure*}[t!]
    \centering
    \includegraphics[width=1\textwidth]{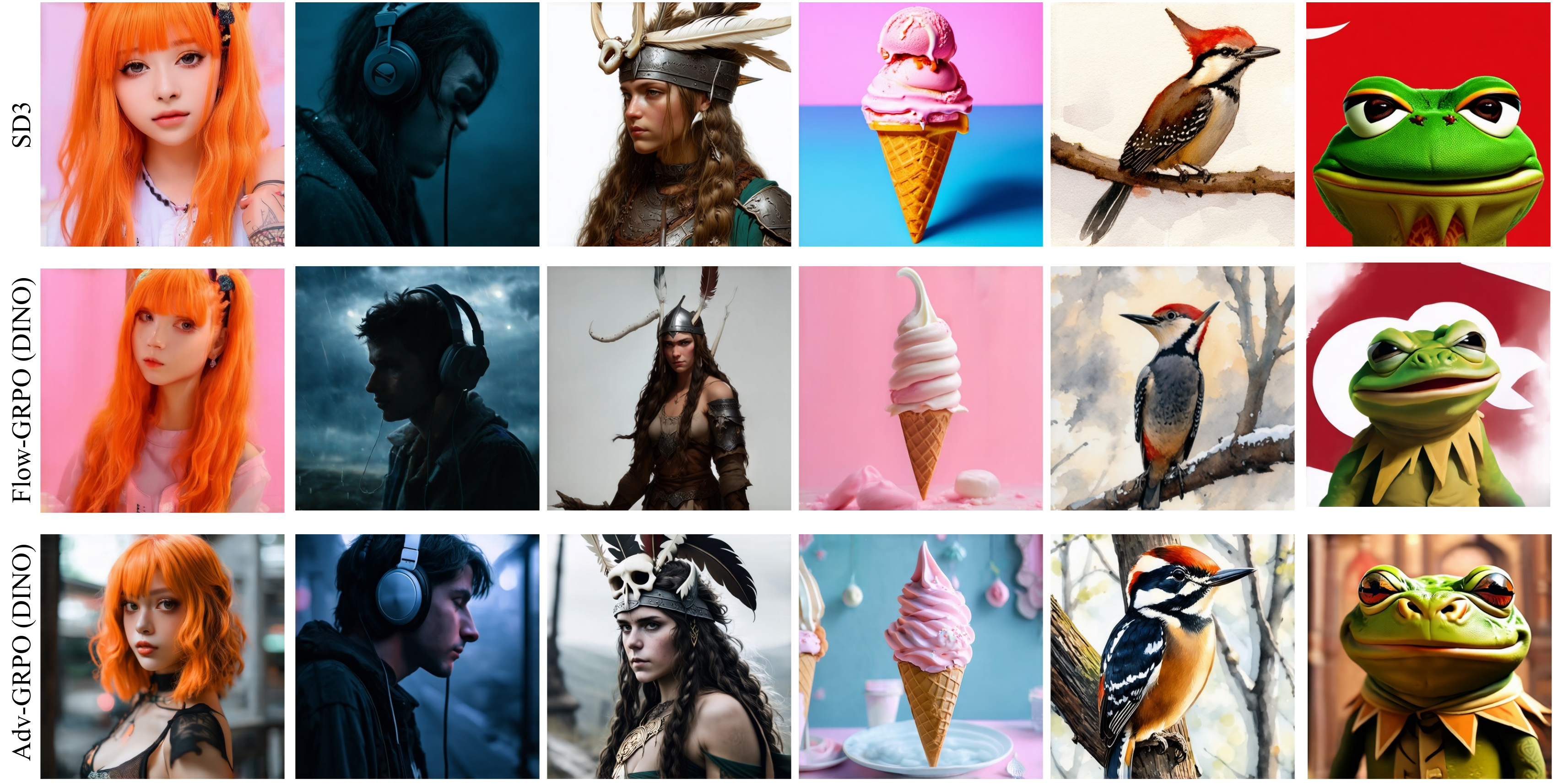}
    \vspace{-20pt}
    \caption{
        Additional visualizations using the DINO reward model. Our method produces images with consistently higher visual quality.
    }
    \vspace{-5pt}
    \label{fig:visualization_dino_more}
\end{figure*}

\noindent\textbf{More Visualizations under DINO reward.} In addition, Fig.~\ref{fig:visualization_dino_more} presents more visualization results obtained under the adversarial DINO reward model. These results show that our method generates images with stronger compositional quality, richer color saturation, improved aesthetic appeal, and more diverse background details, further validating the robustness and generalization ability of our approach.

\section{More Visualizations on Style Customization}
\label{style_transfer}
\begin{figure*}[t!]
    \centering
    \includegraphics[width=1\textwidth]{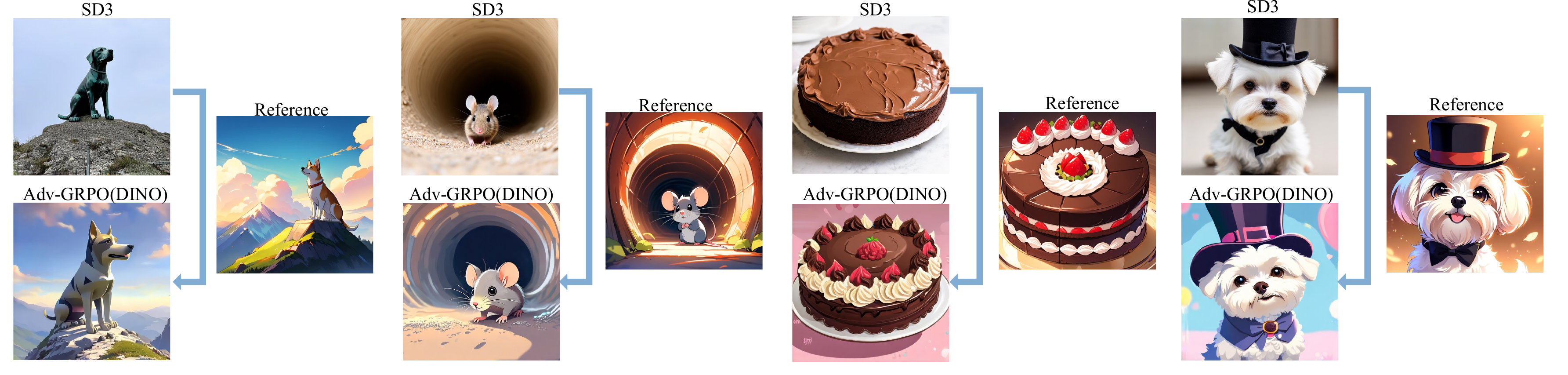}
    \vspace{-25pt}
    \caption{
        More style customization results. Using anime reference images, our method effectively transfers the base model’s style to an anime aesthetic, guided by the provided samples.
    }
    \vspace{-10pt}
    \label{fig:style_transfer_more}
\end{figure*}

As shown in Fig.~\ref{fig:style_transfer_more}, our method successfully transfers the base model’s style to an anime style using anime reference images. These results demonstrate that our RL-based approach, guided by a visual foundation model, can effectively achieve style customization.


\section{Using SigLIP for Optimization}
\label{siglip_optimization}
\begin{figure*}[t!]
    \centering
    \includegraphics[width=1\textwidth]{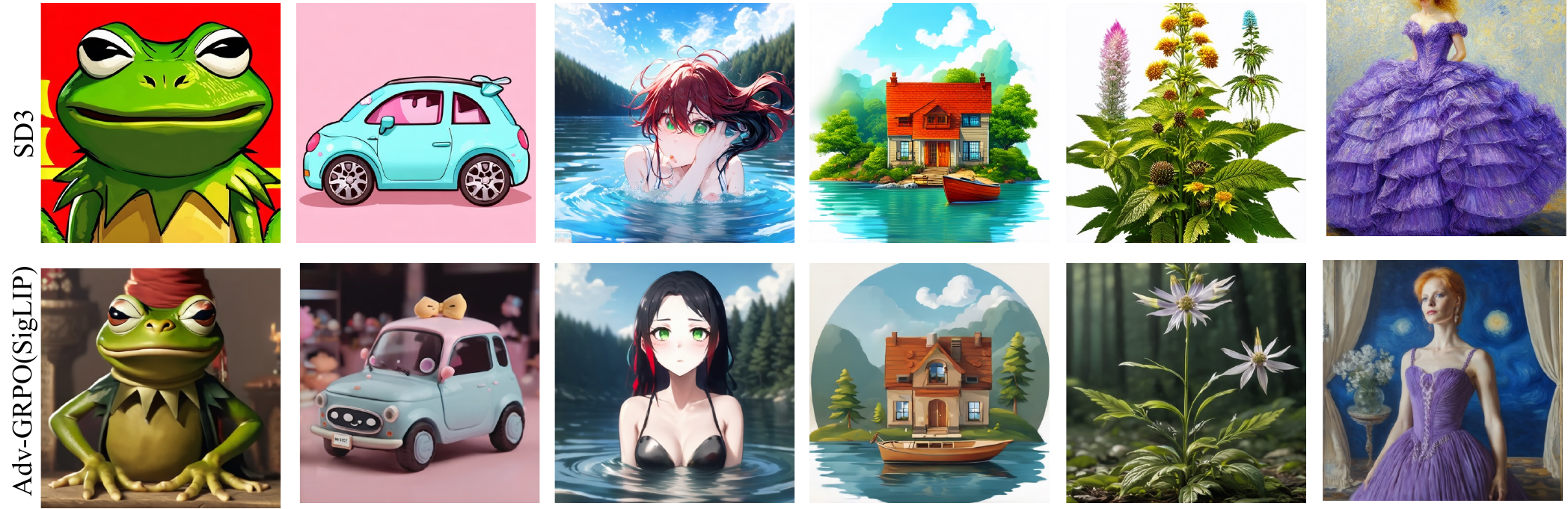}
    \vspace{-20pt}
    \caption{
        Visualizations with the SigLIP reward. Compared with SD3, using other visual foundation models such as SigLIP as the reward function can also lead to overall improvements in image quality.
    }
    \vspace{-10pt}
    \label{fig:siglip_result}
\end{figure*}

As shown in Fig.~\ref{fig:siglip_result}, in addition to DINO, we also experiment with SigLIP as the visual foundation model used for optimization. The pipeline follows the same structure as DINO: we attach a lightweight head to SigLIP and use it to classify reference images and generated images. In this setup, SigLIP serves as the discriminator, while SD3 functions as the generator. Unlike DINO, which provides both global and local features, SigLIP offers only global representations. The successful performance under SigLIP demonstrates that \textbf{our method generalizes well to visual foundation models beyond DINO.}

\begin{figure*}[t!]
    \centering
    \includegraphics[width=1\textwidth]{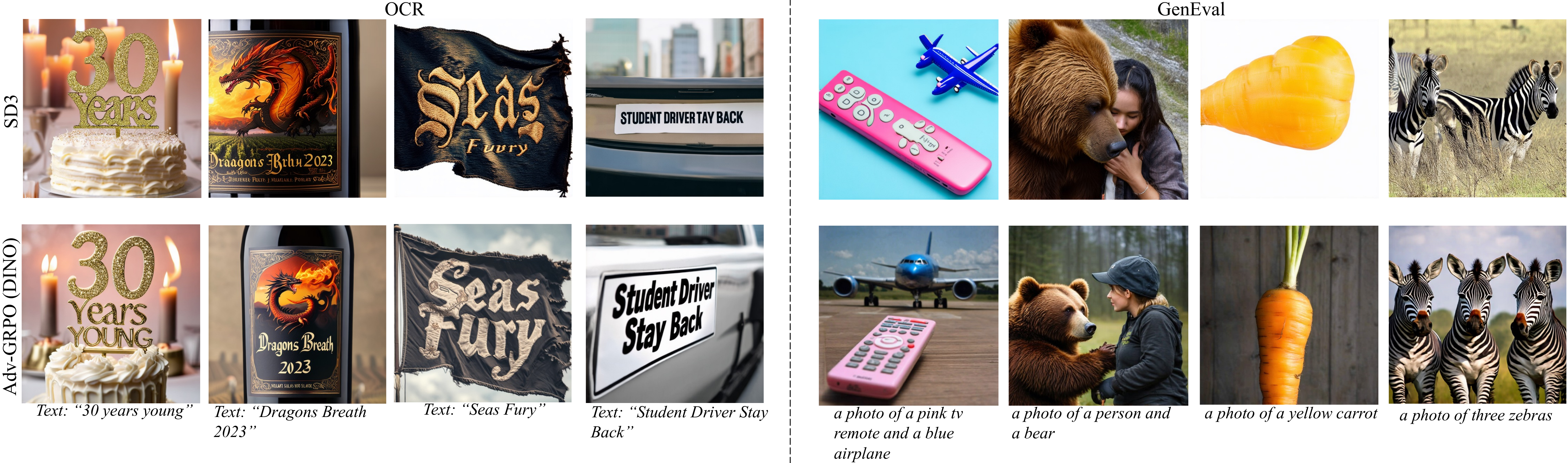}
    \vspace{-20pt}
    \caption{ More visualizations with DINO reward using        different benchmark OCR and GenEval prompts. }
    \vspace{-15pt}
    \label{fig:dino_ocr_geneval_more}
\end{figure*}

\begin{figure*}[t!]
    \centering
    \includegraphics[width=1\textwidth]{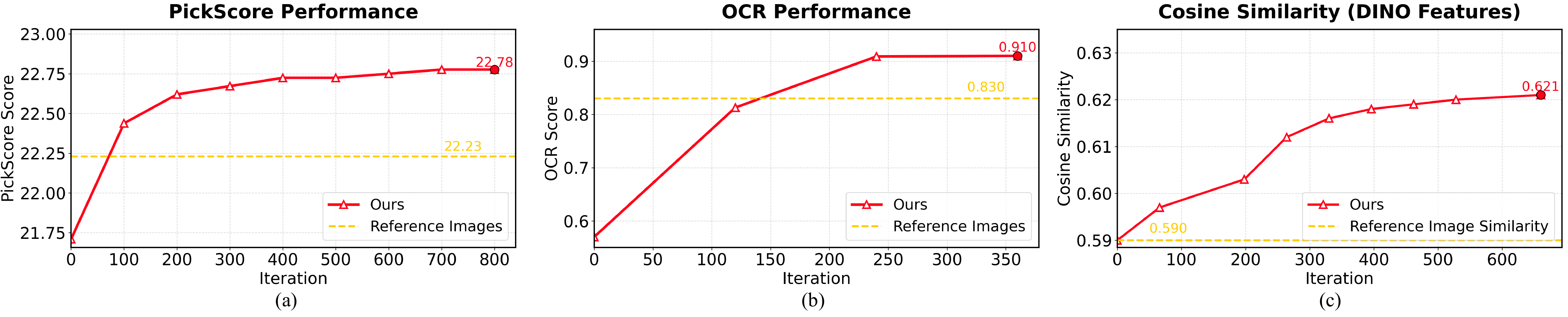}
    \vspace{-15pt}
    \caption{
        \textbf{Reward curves under different reward models.}
        We shows the training dynamics of our method and the baseline under three reward models:
        (a) PickScore, (b) OCR accuracy, and (c) DINO cosine similarity.
    }
    \vspace{-10pt}
    \label{fig:reward_curve}
\end{figure*}


\begin{figure*}[t]
    \centering
    \includegraphics[width=0.9\textwidth]{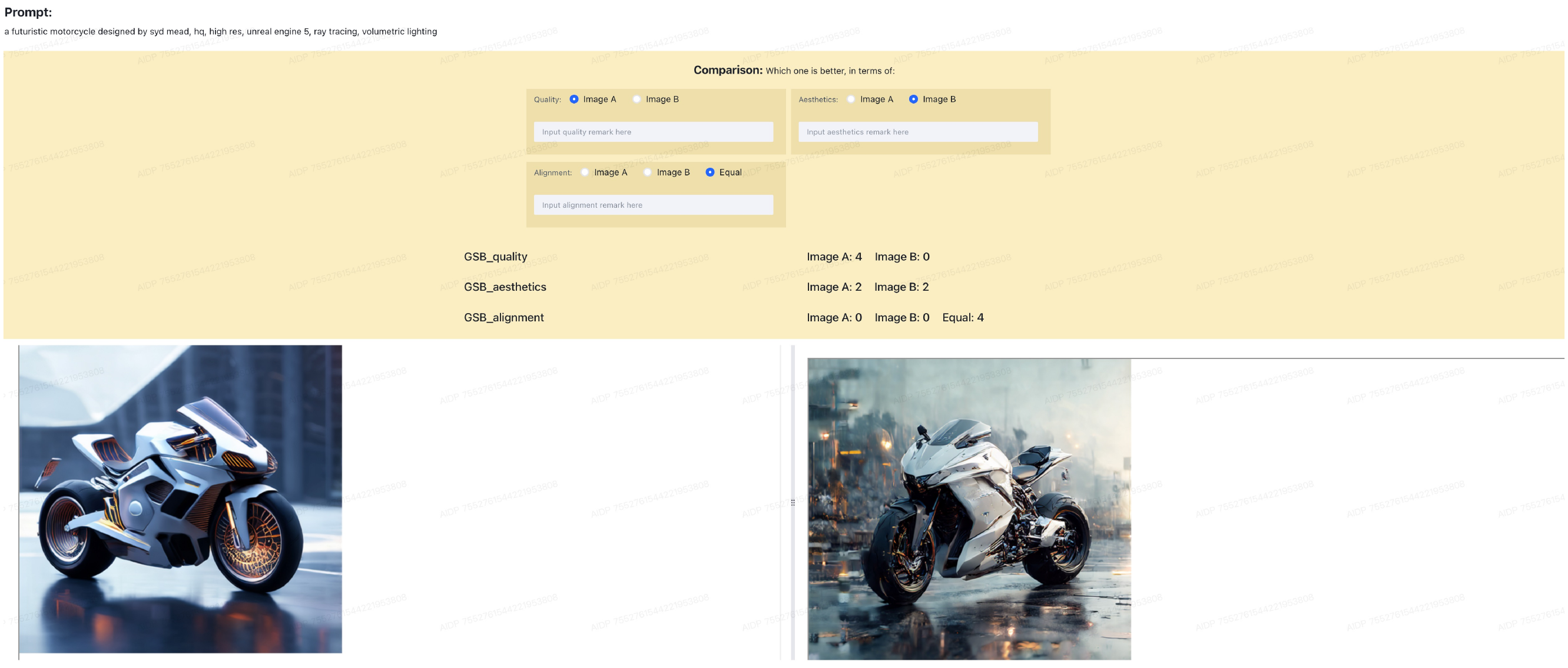}
    \vspace{-10pt}
    \caption{
    Screenshot of the interface used in our human evaluation study.
    }
    \vspace{-15pt}
    \label{fig:human_eval_demo}
\end{figure*}

\section{Reward Curve}
\label{reward_curve}

We report the reward curve obtained during training, as illustrated in Fig.~\ref{fig:reward_curve}. The results show that training converges within approximately 1000 steps. In addition, the reward of our generated images consistently surpasses that of the reference images (produced by the QWen model) throughout the training process.

\section{Human Evaluation}
\label{more_human_evaltion}
For the human evaluation, we assess model performance across three dimensions: \textit{image quality}, \textit{image aesthetics}, and \textit{text-image alignment}. For each question, experts are presented with two images generated by two different models and are asked to select the better one along all three dimensions, as shown in Fig.~\ref{fig:human_eval_demo}.

We construct a benchmark consisting of \textbf{10 groups} of comparison tasks, with a total of \textbf{100 questions}. Each group is evaluated by \textbf{12 experts}, and each question receives annotations from \textbf{3 independent experts}. This setup results in \textbf{300 individual annotation data points} (100 questions $\times$ 3 annotators per question), from which we derive the final aggregated results.

To ensure the reliability of the human evaluation, we adopt a multi-step quality-control protocol. First, we conduct \textbf{expert calibration}, during which annotators review reference examples and align on the scoring criteria. During the evaluation, we monitor and \textbf{resolve inconsistent annotations} through cross-checking and adjudication when needed. In addition, we \textbf{continuously verify and refine the scoring guidelines} throughout the evaluation to minimize ambiguity and ensure consistent interpretation across annotators.

\end{document}